# Content-based similar document image retrieval using fusion of CNN features

Mao Tan
The College of Information Engineering, Xiangtan University, Xiangtan 411105
China
mr.tanmao@gmail.com

Siping Yuan
The College of Information Engineering, Xiangtan University, Xiangtan 411105
China
201610171906@smail.xtu.edu.cn

Yongxin Su
The College of Information Engineering, Xiangtan University, Xiangtan 411105
China
su_yong_xin@163.com

## ABSTRACT

Rapid increase of digitized document give birth to high demand of document image retrieval. While conventional document image retrieval approaches depend on complex OCR-based text recognition and text similarity detection, this paper proposes a new content-based approach, in which more attention is paid to features extraction and fusion. In the proposed approach, multiple features of document images are extracted by different CNN models. After that, the extracted CNN features are reduced and fused into weighted average feature. Finally, the document images are ranked based on feature similarity to a provided query image. Experimental procedure is performed on a group of document images that transformed from academic papers, which contain both English and Chinese document, the results show that the proposed approach has good ability to retrieve document images with similar text content, and the fusion of CNN features can effectively improve the retrieval accuracy.

## 1 INTRODUCTION

Due to development of digital media technology, the scale of multimedia resources including the document images is getting bigger and bigger. Document image retrieval, the task of which is to find useful information or similar document images from a large dataset for a given user query, has become an important research domain in natural language processing. Many approaches based on Optical Character Recognition (OCR) have been proposed, which recognize text content from images and then use text similarity detection to implement document image retrieval system.

Conventional document image retrieval depends on complex model of the OCR-based approach, has some weaknesses such as high computational cost, language dependency, and it is sensitive to image resolution. Direct recommendation and retrieval on the basis of arbitrary multi-character text in unconstrained image require a recognition-free retrieval approach to learn and recognize deep visual features in images. The new document image recognition-free retrieval approach will be conducive to detect the re-contributed and re-published text content on the database of academic journals theses, or query the relevant literature in massive resources.

Document images may be noisy, distorted, and skewed, digitized text need to be processed using different pre-processing methods. According to the type of document image dataset, various pre-processing methods are applied to the document images. In some cases, converting colourful images to grayscale images, adjustment of images' sizes, border removal and normalization of the text line width in the initial steps can enhance document images [1-3].

In early studies on text recognition and retrieval, the extraction of features requires layout analysis, line segmentation, word segmentation, word recognition, etc. But over the last decade, deep learning based features extraction has become an key research direction. Among various deep learning models, the Convolutional Neural Networks (CNNs) are the most powerful networks in image processing tasks. When CNNs are trained in images database, a deep representation of the image is constructed to make object information increasingly explicit along the processing hierarchy [4]. During the CNN feature training phase, Redmon et al. [5] proposed an improved model that inspired by the GoogLeNet model[6] for image classification. They pre-trained the model's convolutional layers on ImageNet dataset for approximately a week, and used the initial convolutional layers of the network to extract features from the image while the fully connected layers to predict the result. Gatys et al. [7] obtained a style representation of an input image and generated results on the basis of the VGGNet, which is a CNN that rivals human performance on a common visual object recognition benchmark task [8]. For learning visual features of multi-character text, Ian et al. [9] proposed a unified approach that integrates the localization, segmentation, and recognition steps via the use of a deep convolutional neural network that operates directly on the image pixels. Hong et al. [10] studied the efficacy of the conceptual relationships by applying them to augment imperfect image tags, and then the relevant results are subsequently used in content-based image retrieval for improving efficiency.

However, compared with other similar methods, the parameter space of CNN network is too large to train a CNN model in a short time. Fortunately, there are some open pre-trained models that we can easily use, such as MatConvNet [11]. Besides that, training the CNN features on a large dataset and fine-tuning by target dataset can significantly improve the performance [12]. Furthermore, we can use the PCA method to reduce the dimension of the CNN features according to the investigation in reference [13], which is mainly to evaluate the performance of compressed neural codes, and it declared that plain PCA or a combination of PCA with discriminative dimensionality reduction can result in very short codes and good (state-of-the-art) performance.

To the best of our knowledge, model fusion is a very powerful technique to increase accuracy on a variety of machine learning tasks. The most basic and convenient way to fusion is to ensemble the features or predictions from multiple different models on the test set, which is a quick way to ensemble already existing model when teaming up. When averaging the outputs from multiple different models, not all predictors are perfectly calibrated or the predictions clutter around a certain range. Fusion methods is key to the solutions, better results can be obtained, if it is given by a linear combination of the ensemble the features or predictions. In this case, the combination coefficients have to be determined by some optimization procedure [14]. A ranking average is proposed in [15] that first turn the predictions into ranks, and then averaging these ranks, which do well on improving the exact same fusion used an average. Moreira et al. [16] specifically tested two un-supervised rank aggregation approaches well known in the information retrieval literature, namely CombSUM and CombMNZ. These algorithms are used to aggregate the information gathered from different outputs or features in order to achieve more accurate ranking results than using individual scores.

Similarity measurement is another key technique to determine the effectiveness of the retrieval system. There are many ways to measure the similarity of image content. An efficient and widespread method is computing pair-wise image cosine similarity based on visual features of all images, and then used this parameter value to retrieve the high similarity images [17].

In this paper, we try to establish a content-based approach to document image retrieval with the purpose of finding out the similar document through a query document image. We choose a document image similarity retrieval method with CNN feature extraction and cosine similarity matching as a basic framework. At the same time, a multiple models fusion method is proposed, which using *Rank_age* of each CNN network to obtain the weighted average fusion feature, and then integrate these methods in the framework in order to improve the accuracy of retrieval system. In the experimental procedure, we slice a batch of English and Chinese academic papers into document images as the image dataset, a group of document images with changed text content is used as the query image, several case studies are provided to evaluate the adaptability and accuracy of the proposed method in different conditions.

## 2 METHODOLOGY

In this section, we mainly discuss several key steps of the document image similarity retrieval based on the multiple CNN models fusion features of images. Firstly, we fine-tune the pre-trained CNN models using MatConvNet, and set repeatedly the crop size of experimental document image. After that, we use multiple different fine-tuned CNN models to extract diverse CNN features from experimental document image dataset, which can convert the visual content into a deep representation. As the CNN feature matrix trained by the CNN model are high-dimensional, we further perform the PCA method to reduce the dimensions and make the each CNN feature matrix has identical dimension in order to subsequent model fusion. Then ensemble the multiple CNN feature matrix by corresponding combination coefficients that calculated from the *Rank_age* of its CNN network, obtain a weighted average fusion feature. After that, we compute and rank the cosine similarity of document images to the query images based on the weighted average fusion feature, output the final retrieval result. In the following section, we elaborate on each of steps in detail, the entire processes are shown in Figure 1.

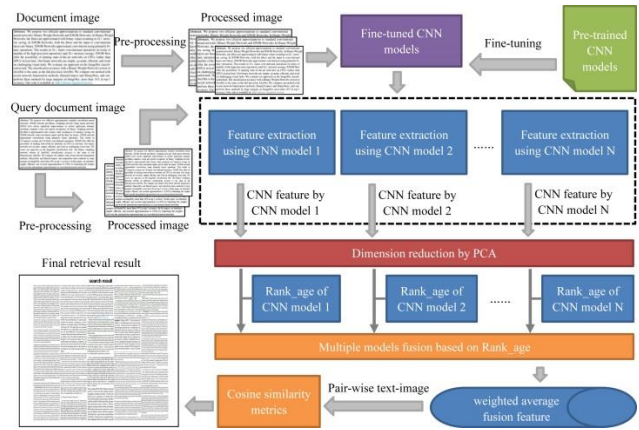

**Figure 1: The entire process of the proposed approach**

As shown in Figure 1, firstly, we convert original images to processed images by using some mature pre-processing methods. We extract the CNN feature to obtain the deep visual representations by fine-tune the multiple pre-trained CNN network model on the target document image dataset. After obtaining the CNN feature matrix, we reduce the dimension of the matrix and improve the efficiency of the algorithm by PCA. Then, we ensemble the various features from multiple network models based on *Rank_age* value. We measure the cosine similarity between the query document image and each image in the training dataset based on multiple model fusion features, and show the most similar images.

### 2.1 CNN Feature Extraction

It is necessary to extract the primitive features of document image as the constructive parameter of the training model. The quality of the feature extraction directly determines the retrieval effect.

Recently, CNNs have achieved impressive results in some areas such as image recognition and object detection. It can input image into the network directly, avoiding the complex feature extraction and data reconstruction process in traditional recognition algorithm. As described above, we choose some state-of-the-art CNN models that submitted for the ImageNet challenge over the last 5 years as the training network. Among them, AlexNet, the first entry that use a deep neural network in 2012 ImageNet competition, has strong generalization ability in computer vision tasks. VGGNet is a preferred multi-layer neural network model for extracting the CNN features of image. The VGGNet use small-size convolution filters and deep network layers, which also has strong generalization ability in many computer vision applications and other image recognition datasets. GoogLeNet can improve utilization of the computing resources inside the network by a carefully crafted design, which allows for increasing the depth and width of the network while keeping the computational budget constant, has good prediction performance in image classification. In addition, ResNet uses a residual learning framework to ease the training of deeper networks but still owes low complexity of the network, and it outperforms the human-level accuracy in the 2015 ImageNet competition.

In general, most CNN models are trained by composing simple linear or non-linear filtering operations, while their implementation need to be trained on large dataset and learned from vast amounts of data. Therefore, we fine-tune the above mentioned pre-trained models on the target document image dataset. At the training phase, we input the fixed-size images that turned by a series of pre-processing to multiple networks and removes the mean. After that, we retain the CNN feature matrix of the penultimate layer of this deep CNN representation, which can be used as a powerful image descriptor applicable to many types of datasets.

## 2.2 Dimension Reduction by PCA

After CNN feature extraction with various CNN model, we obtain some high-dimensional image deep representation. We use the PCA method to compress the CNN feature matrix to 256-D. It reduces some information redundancy, and make the CNN feature matrix has identical dimension to facilitate the subsequent model fusion.

In order to avoid the influence of the sample units, and simplify the calculation of covariance matrix, we use the PCA method to find the 256 largest variation feature vectors in this matrix. Therefore, covariance matrix $C$ can be calculated according to each feature vector $x_i$ in normalized CNN feature matrix, which can be expressed as

$$C = \frac{1}{n}\sum_{i=1}^{n} x_i x_i^T, \quad (1)$$

where $C$ represents the covariance matrix of the feature matrix, and $n$ represents the number of feature vectors.

After that, the eigenvalue equation based on $C$ can be expressed as

$$\lambda_i \mu_i = C \mu_i, \quad (2)$$

where $\lambda_i$ is the eigenvalue of the covariance matrix, and $\mu_i$ is the corresponding eigenvector of the covariance matrix.

Then, we use the resulting 256 normalized feature vectors to constitute the main feature matrix to form a 256-D space. Based on that, we project the high-dimensional CNN feature matrix onto the 256-D dimensional space. Finally, the CNN feature projection matrix is indexed to improve the retrieval efficiency.

## 2.3 Fusion of CNN Features

Through the above method, we obtain various fine-turned CNN models to extract image features. It has been confirmed that creating ensembles from multiple individual files can reduces the generalization error. Therefore, we fuse the features from multiple different existing models respectively, propose the multiple models fusion method based on *Rank_age*.

The features trained by different CNN models might represent different characteristics of document image, and utilizing different features effectively through multiple models fusion method will have positive effect on document image similarity retrieval. We improve the model fusion method that based on ranking average in [15], ensemble the features from multiple model by corresponding combination coefficients that calculated from the *Rank_age* of its model network.

A small scale document image dataset is created in advance to calculate the *Rank_age* of each model, which include 422 pair similar document images and the index of each pair of images. Then, the *Rank_age* of each model can be calculated according to the retrieval results that learned by corresponding model on this dataset, which is more adaptable to ensemble different models that have significant difference. The *Rank_age* can be calculated as

$$Rank\_age = \sum_{i=1}^{n} \frac{score}{rank_i}, \quad (3)$$

where $n$=422, *score* is the mean accuracy in the top-5 similar images to the each query when using certain model, $rank_i$ is the ranking of the $i$-th image's similar image in its retrieval result.

After that, normalizing the *Rank_age* between 0 and 1 can get the corresponding combination coefficient $\varepsilon$. Finally, we ensemble the three CNN feature matrix $M_{VD}$, $M_{VE}$ and $M_G$ trained by VGGNet-D, VGGNet-E and GoogLeNet respectively according to corresponding $\varepsilon$ and obtain the weighted average fusion feature that can be expressed as

$$M = \varepsilon_{VD} * M_{VD} + \varepsilon_{VE} * M_{VE} + \varepsilon_G * M_G, \quad (4)$$

where $\varepsilon_{VD}$, $\varepsilon_{VE}$ and $\varepsilon_G$ are the corresponding coefficients for the feature matrix to ensemble, and $\varepsilon_{VD} + \varepsilon_{VE} + \varepsilon_G = 1$.

## 2.4 Similarity Metric

Cosine similarity has been proved to be an effective metric system because of its accuracy. The 256-D weighted average fusion feature matrix $[z_1, z_2, …, z_n]^T$ could describe the main CNN features of the document images in the dataset, where $n$ is the number of document images in the datasets. The cosine similarity calculated from CNN feature vector can approximately measure the similarity between document images.

For each pair of feature vector ($Z_u$, $Z_v$) where $u \neq v$, the pair-wise image cosine similarity $T_s$ can be expressed as

$$T_s(Z_u, Z_v) = \frac{\sum_{i=1}^{k} F(Z_u, u_i) * F(Z_v, v_i)}{\sqrt{\sum_{i=1}^{k} F(Z_u, u_i)^2} * \sqrt{\sum_{i=1}^{k} F(Z_v, v_i)^2}}, \quad (5)$$

where $K = 256$, and $F(Z_u, u_i)$ is the value of the $i$-th column element of the 256-D dimensional feature vector corresponding to the document image $Z_u$. $T_s(Z_u, Z_v)$ is the pair-wise document image cosine similarity. Through equation (5), we can retrieve out some high similarity document images to query image.

## 3 EXPERIMENTS

### 3.1 Data Collection and Evaluation Metric

In this work, to evaluate the proposed method, we collect a group of English and Chinese academic papers as the text database, and cut them into many small pieces of heterogeneous document image to construct a training dataset, which contains 2017 images totally. Then, we select some text paragraphs from the original article and edit them by various ways. After that, we store the edited text paragraphs as images to construct an query image dataset including 422 images totally, which is used to evaluate the accuracy of the proposed approach in various situations. In addition, we select the 422 query images and their original images to construct a small scale document image dataset, and create a <query image name, original image name> index to calculate the *Rank_age* of each CNN network in advance.

We performed different experiments with different CNN model. 422 query document images is selected to retrieve out the similar document images in image dataset. The proposed method is evaluated using the accuracy value measured based on the results ranked among the Top-1, Top-3, Top-5 and Top-10 similar images to a query document image.

### 3.2 Experimental Results and Analyses

The training and query dataset includes English and Chinese document images, and there are 10 types edited images in query dataset, including retranslating by Google, changing the font color, adding another statement in the content, omitting lots of content, adjusting the line spacing of the text and reversing the word order, and so on. Therefore, we retrieve separately different images each time to see the retrieval effect of different text language and content, or local deformation of layout. At first, we choose an English document image as query image, which is converted from the abstract of the an English article, The query document image is shown as Figure 2(a). After that, we calculate the query document image's similarity to each document image in training dataset by using MMF (VGGNet-D + VGGNet-E + GoogLeNet), the original document image that is shown in Figure 2(b) can be retrieved out in first.

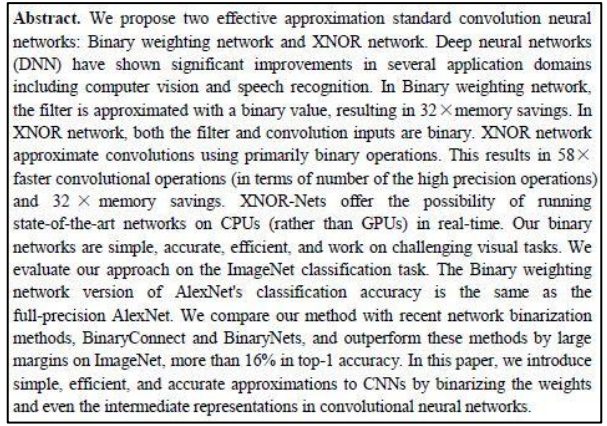

(a)

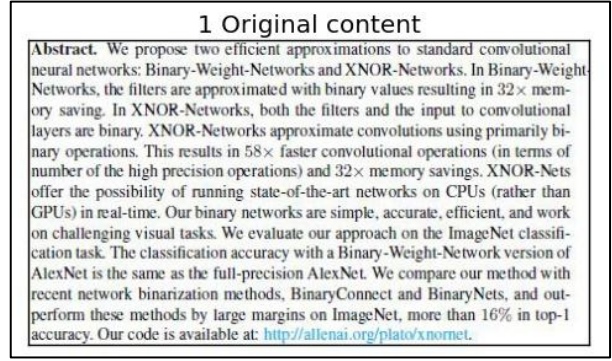

(b)

**Figure 2: The Top-1 similarity retrieval of English document image. (a) Query image. (b) Result image.**

Another case study is provided to evaluate the retrieval effect when querying through Chinese document image, which is re-translated by Google and modified in its original text content. The Top-1 retrieval result image is shown in Figure 3, and in these result we can see that similar text content with some different characters and visual presentations can be recognized by the proposed approach.

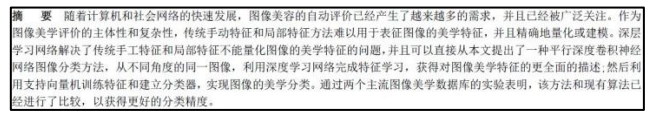

(a)

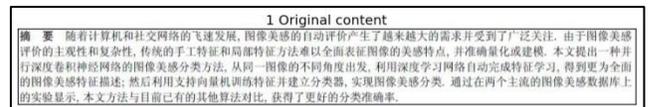

(b)

**Figure 3: The Top-1 similarity retrieval of Chinese document image. (a) Query image. (b) Result image**

Then we consider the Top-1, Top-3, Top-5 and Top-10 accuracy in ranked result using various individual CNN model, and compare them with the accuracy that obtained by multiple models using weighted average fusion feature. The condition that fuse the features of AlexNet and VGGNet-E is named MMF-1, the fusion of AlexNet, VGGNet-D and VGGNet-E is named MMF-2, the fusion of AlexNet and GoogLeNet is named MMF-3, and MMF-4 represents fusion of AlexNet and ResNet-152. According to experimental performance, for GoogLeNet and ResNet, the crop size of document images in the case is fixed as 288×288, and for other models it is set to 256×256. The accuracies obtained from the above situation are shown in Table 1. During the model fusion, we got the *Rank_age* and $\varepsilon$ of each CNN network in advance, which is obtained by training each network with the small scale document image dataset.

**Table 1: The Retrieval Accuracy of Various Models**

| Methods | Top-1 (%) | Top-3 (%) | Top-5 (%) | Top-10 (%) |
|---|---|---|---|---|
| AlexNet | 49.49 | 60.71 | 71.43 | 83.16 |
| VGGNet-D | 34.18 | 44.90 | 53.57 | 63.27 |
| VGGNet-E | 44.90 | 63.78 | 70.92 | 91.33 |
| GoogLeNet | 7.65 | 14.29 | 20.41 | 25.51 |
| ResNet-152 | 29.59 | 42.35 | 49.49 | 64.80 |
| MMF-1 | 50.00 | **75.00** | **86.22** | **97.96** |
| MMF-2 | **52.55** | **75.00** | 85.71 | 96.94 |
| MMF-3 | 49.49 | 61.22 | 71.92 | 83.67 |
| MMF-4 | 49.49 | 60.71 | 70.92 | 83.16 |

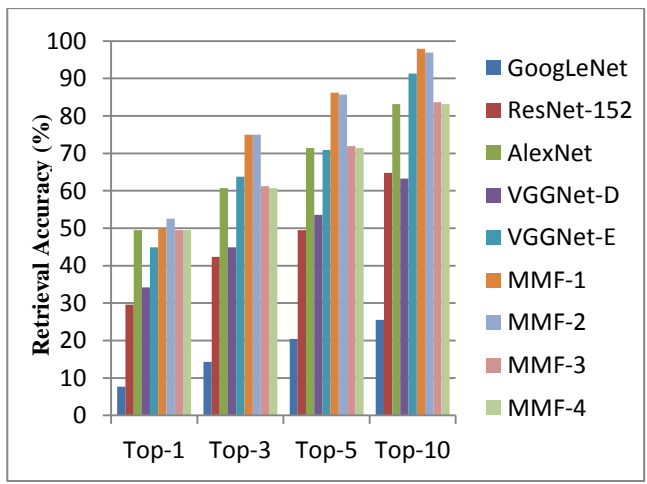

**Figure 4: The retrieval accuracy obtained from the proposed fusion method and some single model.**

In Table 1, it can be seen that by features fusion, as a whole the retrieval accuracy are improved, MMF-1 and MMF-2 obtains better performance on retrieval accuracy than the best individual model AlexNet in our case, which result is mainly caused by good individual models. At the same time, it should be noted that the accuracy improvement fluctuate slightly if GoogLeNet or ResNet-152 is adopted in features fusion. We can see that minor performance improvement in MMF-3 but slight accuracy decreases of MMF-4 in TOP-5 retrieval, which should be caused by the low accuracy of individual model in the two fusion models. The above results show us that the chosen of individual CNN model with good performance is important for the proposed fusion approach. Further in more, Figure 4 illustrates intuitively that the similarity retrieval results through various individual models and the proposed fusion method.

## 4 CONCLUSIONS

In this paper, a new content-based approach to document image retrieval is proposed. All of the experimental results indicate that the proposed approach is effective to realize document image recognition-free retrieval for different language characters without using OCR. By using *Rank_age* to fuse the features obtained from several classical CNN model, the retrieval accuracy can be significantly improved in most of conditions with different transformations of text content or layout. In our next works, for obtaining higher retrieval accuracy, more methods will be chosen and tested to fuse multiple CNN models. When this approach is further improved to adapt more complex transformations, it is expected to be applied in paper plagiarism identification or literature recommendation.

## REFERENCES

[1] Gatos B, Pratikakis I. 2009, Segmentation-free word spotting in historical printed documents. Document Analysis and Recognition, 2009. ICDAR'09. 10th International Conference on. IEEE, 271-275.

[2] Frinken V, Fischer A, Manmatha R, et al. 2012, A novel word spotting method based on recurrent neural networks. IEEE transactions on pattern analysis and machine intelligence, 34(2): 211-224.

[3] Hong R, Zhang L, Tao D. 2016, Unified photo enhancement by discovering aesthetic communities from flickr. IEEE transactions on Image Processing 25(3): 1124-1135.

[4] Gatys, L. A., Ecker, A. S. & Bethge, M. 2015, Texture synthesis and the controlled generation of natural stimuli using convolutional neural networks. arXiv:1505.07376, http://arxiv.org/abs/1505.07376.

[5] Redmon J, Divvala S, Girshick R, et al. You Only Look Once: Unified, Real-Time Object Detection. Computer Science, 2016:779-788.

[6] Szegedy C, Liu W, Jia Y, et al. 2015, Going deeper with convolutions. Proceedings of the IEEE Conference on Computer Vision and Pattern Recognition, 1-9.

[7] Gatys L A, Ecker A S, Bethge M. 2015, A Neural Algorithm of Artistic Style. Computer Science.

[8] Simonyan, K. & Zisserman, A. Very Deep Convolutional Networks for Large-Scale Image Recognition. arXiv:1409.1556, http://arxiv.org/abs/1409. 1556.

[9] Goodfellow I J, Bulatov Y, Ibarz J, et al. 2013, Multi-digit Number Recognition from Street View Imagery using Deep Convolutional Neural Networks. Computer Science.

[10] Vedaldi A, Lenc K. 2014, MatConvNet: Convolutional Neural Networks for MATLAB. 689-692.

[11] Hong R, Yang Y, Wang M, Hua XS. 2015, Learning visual semantic relationships for efficient visual retrieval. IEEE Transactions on Big Data 1(4): 152-161.

[12] Chatfield K, Simonyan K, Vedaldi A, et al. 2014, Return of the Devil in the Details: Delving Deep into Convolutional Nets. Computer Science.

[13] Babenko A, Slesarev A, Chigorin A, et al. 2014, Neural Codes for Image Retrieval, 8689:584-599.

[14] Jahrer M, Töscher A, Legenstein R. 2010, Combining predictions for accurate recommender systems. Proceedings of the 16th ACM SIGKDD international conference on Knowledge discovery and data mining. ACM, 693-702.

[15] kaggle ensembling guide. (2015-6-11). https://mlwave.com/kaggle-ensembling-guide/

[16] Moreira C, Martins B, Calado P. 2015, Using rank aggregation for expert search in academic digital libraries. arXiv:1501.05140, http://arxiv.org/abs/1501.05140.

[17] Sejal D, Rashmi V, Venugopal K R. 2016, Image recommendation based on keyword relevance using absorbing Markov chain and image features. International Journal of Multimedia Information Retrieval, 5(3):1-15.